# Learning Perceptual Concepts by Bootstrapping from Human Queries

Andreea Bobu[1], Chris Paxton[2], Wei Yang[2], Balakumar Sundaralingam[2],
Yu-Wei Chao[2], Maya Cakmak[2,3], and Dieter Fox[2,3]

*Abstract*—When robots operate in human environments, it's critical that humans can quickly teach them new *concepts*: object-centric properties of the environment that they care about (e.g. objects *near*, *upright*, etc). However, teaching a new perceptual concept from high-dimensional robot sensor data (e.g. point clouds) is demanding, requiring an unrealistic amount of human labels. To address this, we propose a framework called Perceptual Concept Bootstrapping (PCB). First, we leverage the inherently lower-dimensional *privileged information*, e.g., object poses and bounding boxes, available from a simulator only at training time to rapidly learn a low-dimensional, geometric concept from minimal human input. Second, we treat this low-dimensional concept as an automatic labeler to synthesize a large-scale high-dimensional data set with the simulator. With these two key ideas, PCB alleviates human label burden while still learning perceptual concepts that work with real sensor input where no privileged information is available. We evaluate PCB for learning spatial concepts that describe object state or multi-object relationships, and show it achieves superior performance compared to baseline methods. We also demonstrate the utility of the learned concepts in motion planning tasks on a 7-DoF Franka Panda robot.

*Index Terms*—Human-Centered Robotics, Human Factors and Human-in-the-Loop, Visual Learning

## I. INTRODUCTION

**R**OBOTS are increasingly expected to perform tasks in human environments, from helping with household chores to cleaning up the office. To align robot performance with the end user's unique needs, the person should be able to teach their robot a new *concept*: an object-centric property of the environment that they care about. A concept maps the environment state to a value indicating how much the object-centric property is expressed, and the robot can optimize it to perform the person's desired task. For example, in Fig. 1 the user wants the robot to tidy the tabletop by moving the mug near the can. To accomplish this behavior, the robot first learns the concept of what it means for objects to be *near* each other, and then moves the mug closer to the can.

To be usable in the real world, a learned concept must operate on an input space the robot understands: the high-dimensional observations from its sensors (e.g. point clouds). Classical methods simplify the perceptual concept learning

Manuscript received: February, 24, 2022; Revised May, 13, 2022; Accepted June, 14, 2022. This paper was recommended for publication by Editor Dana Kulic upon evaluation of the Associate Editor and Reviewers' comments. We thank Weiyu Liu for an aligned version of the Shapenet dataset.
[1] EECS at UC Berkeley abobu@berkeley.edu
[2] NVIDIA Robotics, USA {cpaxton, weiy, balakumars, ychao, dieterf}@nvidia.com
[3] University of Washington mcakmak@cs.washington.edu
Digital Object Identifier (DOI): see top of this page.

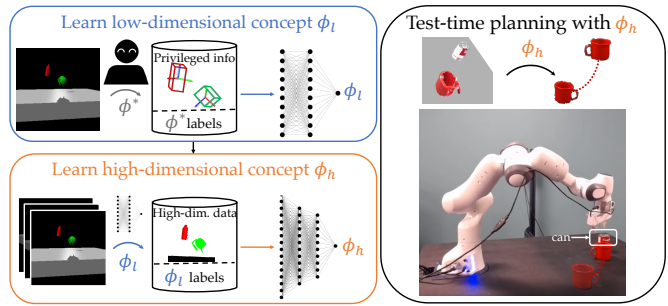

Fig. 1: (Left) We propose a new approach whereby the robot collects labels from the human about the concept $\phi^*$ (e.g. *near*) and learns a low-dimensional concept $\phi_l$ on the *privileged information* space (e.g. poses and bounding boxes) (top), then uses $\phi_l$ to label data necessary for learning the high-dimensional concept $\phi_h$ (bottom). (Right) At test time, the robot can directly use the learned concept $\phi_h$ to produce a plan for moving the mug to be *near* the can. Additional qualitative results available at https://sites.google.com/nvidia.com/active-concept-learning.

problem via a pre-processing step, extracting geometric information like object poses and bounding boxes from the high-dimensional sensor data [1], [2], [3]. The human can then teach a concept mapping from this lower dimensional space by labeling when objects are near or far. By transforming the sensor input into a lower dimensional space, the robot can learn the concept quickly even from limited human input. Unfortunately, recovering accurate geometries from real-world sensor data is challenging: even modern pose estimators [4], [5], [6], [7] struggle when confronted with partial occlusions or novel objects [8]. Instead, recent deep learning alternatives learn perceptual concepts directly from the sensor data, without any pre-processing [9], [10]. These methods are usually trained in simulation, where a variety of objects can be manipulated in diverse configurations, resulting in better generalization than classical approaches [8], [11], [10]. However, because of the high dimensionality of the input space, the robot needs unreasonably many human-labeled examples, making a new concept impractical and cumbersome for a human to teach.

In this paper, we propose getting the best of both worlds: learn concepts from high-dimensional sensor data with limited human labeling effort. We observe that, while the robot only has access to the high-dimensional sensor inputs during task execution, at training time it has a simulator containing *privileged information* akin to the geometries that classical approaches tried to compute. In Fig. 1, this privileged space is the object poses and bounding boxes – a much simpler representation than their high-dimensional point cloud equivalent. Our idea is to learn a low-dimensional, geometric variant of the concept on the privileged space, then treat it as a labeler and use it to automatically label high-dimensional data in the simulator (Fig. 1). This lets us generate a large, diverse, and *automatically labeled* dataset for training a high-dimensional, perceptual concept which can be directly applied to real-world settings without additional human input.

Since these low-dimensional spaces result in faster training and are often semantically meaningful, they allow for richer human interaction. We thus investigate three types of human input for concept learning (demonstration, label, and feature queries), and propose an active learning strategy for informative queries. We then showcase PCB in experiments both in simulation and on a real Franka Panda robot.

## II. RELATED WORK

**Concept learning from low-dimensional geometries.** Traditionally, concepts are hand-engineered by the system designer prior to robot deployment [12], [13], [14]. Unfortunately, by relying entirely on prior specification, the robot cannot adapt its task execution to an end user's needs. Recent works address this problem by allowing the robot to either infer concepts from task demonstrations [15], [16] or learn them directly from the human [17], [18]. While these methods enable the robot to learn after deployment, they have been primarily demonstrated on low-dimensional spaces.

Prior work bypasses the high-dimensional learning problem by extracting low-dimensional geometric information and learning relational concepts on top of it [1], [2], [3]. However, recovering accurate geometries from real-world sensor data is challenging: even modern pose estimators [4], [5], [6], [7] struggle with the partial occlusions or novel objects that appear in open-world environments [8]. As such, we seek to learn concepts operating directly on high-dimensional input, without any intermediary pre-processing.

**Learning from high-dimensional sensor data.** Deep learning handles high-dimensional data by using a function approximator to learn low-dimensional embeddings, hoping to capture salient aspects of the environment. Deep inverse reinforcement learning (IRL) and imitation learning approaches, in particular, use demonstrations to automatically extract behavior-relevant representations [19], [20], [21]. Unfortunately, to work reliably on high-dimensional inputs, these methods require a large amount of data from the human to generalize outside of the training distribution [22], [23].

Recent work in the auto-encoder community suggests that we can improve data efficiency by learning a disentangled latent space from weakly labeled examples of many concepts [24]. Unfortunately, this approach still requires the user to label tens of thousands of examples for training. Moreover, while these methods are aligned with our goals of capturing important aspects of robotic tasks, they are complementary in that they focus on learning latent embeddings of the high-dimensional data, not individual perceptual concepts.

**Concept learning from high-dimensional sensor data.** Instead of learning a universal representation, other work learns specific relational concepts directly from high-dimensional data [9], [10]. In particular, these methods learn from segmented object point clouds, which are easy to obtain and have successfully been used in other perception pipelines [25]. The disadvantage of this approach is that it still requires large amounts of data (thousands of labeled examples), making it unsuitable for learning the concept from a human. We look at how we can teach similar perceptual concepts from high-dimensional point cloud data, but do so quickly and efficiently with the help of the privileged space.

## III. METHOD

Our goal is to enable humans to teach robots perceptual concepts operating in high-dimensional input spaces, like segmented object point clouds. We assume that the robot may query the human for labeled examples of the desired concept, but we wish to learn the concept with as few human labels as possible. As training high-dimensional concepts is data intensive [9], [10], we propose to learn the concept first in a simpler, lower dimensional input space, then use it to label as much high-dimensional data as needed to train the concept in the target high-dimensional sensor space.

### A. Preliminaries

Formally, a concept is a function mapping from input state to a scalar value, $\phi(s) : \mathbb{R}^d \to [0, 1]$, indicating how much concept $\phi$ is expressed at $d$-dimensional state $s \in \mathbb{R}^d$. In our setting, we assume the human teacher already knows the ground truth concept $\phi^*$ and can answer queries about it.

At training, the robot has a simulator that gives it access to the entire state $s$, but at test time it receives high-dimensional observations $o_h \in \mathbb{R}^h$ given by a transformation of the state $\mathcal{F}(s) : \mathbb{R}^d \to \mathbb{R}^h$. In the example in Fig. 1, $s$ captures the objects' pose, mesh, color, etc, whereas $o_h$ is only the segmented point cloud of the scene from a fixed camera view. The robot seeks to learn a high-dimensional concept mapping from these observations, $\phi_h(o_h) : \mathbb{R}^h \to [0, 1]$, so that it can use it in desired manipulation tasks later on.

To do so, we assume the robot can ask the person for state-label examples $(s, \phi^*(s))$, forming a data set $\{s, o_h, \phi^*(s)\} \in \mathcal{D}_\phi$. Since the high-dimensional observation $o_h$ directly corresponds to state $s$, this data set has the crucial property that the same label $\phi^*(s)$ applies to both $s$ and $o_h$:

$$\phi^*(s) = \phi_h(o_h), \forall s, o_h = \mathcal{F}(s) \ . \tag{1}$$

From here, one natural idea to learn $\phi_h$ is to treat it as a classification or regression problem and directly perform supervised learning on $(o_h, \phi^*(s))$ pairs. Unfortunately, to learn a meaningful decision boundary, this approach would require very large amounts of data from the person, making it impractical to have a user teach a new concept.

Instead, we assume the robot can use *privileged information* from the simulator as a low-dimensional observation $o_l \in \mathbb{R}^l$ given by a transformation $\mathcal{G}(s) : \mathbb{R}^d \to \mathbb{R}^l$. We think of this information as privileged because the robot has access to it during training but not at task execution time. In Fig. 1, $o_l$ only needs the object poses and bounding boxes to determine whether the objects are near. The set of collected human examples then includes the low-dimensional observation: $\{s, o_l, o_h, \phi^*(s)\} \in \mathcal{D}_\phi$, which allows the robot to learn a low-dimensional variant of the concept, $\phi_l(o_l) : \mathbb{R}^l \to [0, 1]$, by extending the property in Eq. (1):

$$\phi^*(s) = \phi_h(o_h) = \phi_l(o_l), \forall s, o_h = \mathcal{F}(s), o_l = \mathcal{G}(s) \ . \quad (2)$$

We hypothesize that learning the low-dimensional concept $\phi_l$ from privileged information should require less human input than learning $\phi_h$ directly from high-dimensional data. Moreover, Eq. 2 allows the learned $\phi_l$ to act as a labeler, bypassing the need for additional human input. We, thus, break down the concept learning problem into two steps: leverage the human queries to learn a low-dimensional concept $\phi_l$, then use it to ultimately learn the original high-dimensional $\phi_h$.

### B. Learning a low-dimensional concept

To learn $\phi_l$ the robot first needs to ask the human for $\mathcal{D}_\phi$. To ensure the robot can learn the concept with little data, we want a query collection strategy that balances being informative and not placing too much burden on the human. We consider two types of input that are easy to provide and commonly used in the human-robot interaction (HRI) literature [26]: *demonstration queries* and *label queries*. Since users may struggle to label continuous values, we simplify the labeling to consist of 0 (negative) or 1 (positive) for low and high concept values. Note that despite the labels being discrete, they can still be used to learn a model that predicts continuous values.

*Demonstration queries*, or *demo queries*, involve creating a new scenario and asking the human to choose states $s$ that demonstrate the concept and label them according to $\phi^*$. This method requires an interface that allows the person to directly manipulate the state of the environment and label it, like a simulator with keyboard or click control. For example, for the *near* concept in Fig. 1, the person could move the red object near the green one and label the state 1.0, symbolizing a high concept value. Here, the robot can only manipulate the constraints of the scenario (e.g. which objects are involved) and the human has complete control over the selection of the rest of the state (e.g. the objects' poses).

If the human is pedagogic, demonstration queries provide the robot with an informative data set of examples that should allow it to learn the low-dimensional concept quickly. Unfortunately, this data collection method can be quite slow due to the fact that the person has to spend time both deciding on an informative state and manipulating the environment to reach it. This makes it challenging to use in data intensive regimes (like when training $\phi_h$ from the get-go) but ideal in the low-data ones we are interested in.

*Label queries* are a less time-consuming alternative where the robot synthesizes the full query state $s$, and the person simply labels it as 0 or 1. For instance, in Fig. 1 the robot picks both the objects and their poses. This type of query is much easier and faster for the person to answer, but places the burden of informative state generation entirely on the robot. Simply randomly sampling the state space might not be very informative for the desired concept. For example, for a concept like *above*, placing two objects at random locations will rarely result in examples where one object is above the other. As such, we need a way to select more useful queries.

We use active learning [27], [28], whereby the robot can proactively select query states that it deems more informative. Concretely, we interleave asking for a batch of query states with learning the concept $\phi_l^t$ from the $t$ examples received so far. This way, the robot can use the partially-learned $\phi_l^t$ to inform the synthesis of a more useful next batch of queries. For every query, the robot chooses among three query synthesis strategies: 1. *random*: randomly generate a state $s \in \mathbb{R}^d$; 2. *confusion*: pick the state that maximizes confusion by being at $\phi_l^t(s)$'s decision boundary, i.e. $s = \arg\min_s(\|\phi_l^t(s) - 0.5\|)$; 3. *augment*: select a state that was previously labeled as a positive (or negative, whichever is rarer) and add noise to it. A *random* query serves as a proxy for exploring novel areas of the state space. In a simulator, this query can be generated by randomizing the parameters of the state (e.g. object meshes, poses, etc). The *confusion* query is a proxy for disambiguating areas of the state space where the current concept $\phi_l^t$ cannot determine whether the state has a positive or a negative concept value. The query state in this case is selected by optimizing the concept value to be 0.5 using the cross-entropy method [29], [30]. The *augment* query is useful for concepts where positives (or negatives) are rarer, like in the *above* example.

Active learning is possible when learning low-dimensional concepts because they have much shorter training cycles than their high-dimensional variants. Another advantage of low-dimensional spaces is that, while the transformation $\mathcal{F}$ cannot be modified because the robot is constrained to operate on $o_h$ at test time, we have more flexibility over what $\mathcal{G}$ and $o_l$ can be. We exploit this with a third type of human input called *feature queries* [26]. Feature queries typically involve asking the person whether an input space feature is important or relevant for the target concept. However, this query is only useful if the feature itself is meaningful to the human. We adapt feature queries and ask the person a few intuitive questions about the concept such that the answer informs the choice of the transformation $\mathcal{G}$. For example, a negative answer to the question "Does the size of the objects matter?" lets the robot know that $o_l$ does not benefit from containing object bounding box information. These queries lead to an appropriate $\mathcal{G}$, which can further speed up the learning of $\phi_l$.

Given a (possibly partial) dataset of labeled human examples $\mathcal{D}_\phi$, the robot can now train a low-dimensional concept $\phi_l$. We treat concept learning as a classification problem, approximate $\phi_l$ by a neural network, and train it on the $(o_l, \phi^*(s))$ examples in $\mathcal{D}_\phi$ using a binary cross-entropy loss.

### C. Learning a high-dimensional concept

Learning a high-dimensional concept requires a large amount of labeled high-dimensional data. Generating this data set is a two-step process: the robot needs to synthesize a large and diverse set of states $s$, which it then has to acquire labels for. However, as opposed to the low-dimensional case, this data set need not be directly labeled by the human: the learned low-dimensional concept itself can act as a labeler.

Since at training time the robot has access to the simulator, for the data synthesis step we randomly explore the state space. With the property in Eq. (2), we can use the low-dimensional concept $\phi_l$ to automatically label the states, generating the data set $\{s, o_l, o_h, \phi_l(o_l)\} \in \mathcal{D}_{\phi_l}$. To now learn the high-dimensional concept $\phi_h$, we approximate it by a neural network and train it via classification on the $(o_h, \phi_l(o_l))$ examples in $\mathcal{D}_{\phi_l}$ using a cross-entropy loss.

### D. Implementation details

We used a multilayer perceptron (3 layers, 256 units) and a standard PointNet [31], [32] to represent the low- and high-dimensional concepts, respectively. Our concepts involved relationships between objects, so we represented the high-dimensional observation $o_h$ with the relevant objects' segmented point clouds from the camera view, and the low-dimensional one $o_l$ with object poses and bounding boxes.

For data generation, we modified the objects in the ShapeNet data set [33] such that they are consistently aligned and scaled. When synthesizing states $s \in \mathbb{R}^d$, we spawned pairs of two objects in the Isaac Gym simulator [34] and manipulated their poses, as well as the camera pose along the table plane. This process resulted in a variety of states with possibly occluded objects, from many camera views. Since our method allows us to generate as much simulated data as desired, our hope is to generalize to real-world conditions like other simulation-based methods do [9], [35], [36].

## IV. EXPERIMENT: LEARNING PERCEPTUAL CONCEPTS BY BOOTSTRAPPING FROM HUMAN QUERIES

In this section, we compare our label-efficient perceptual concept learning method PCB to a baseline that learns directly from high-dimensional input. PCB relies on a human-trained low-dimensional concept $\phi_l$, for which we perform an extensive investigation in Sec. V.

### A. Experimental Design

Throughout our experiments, we synthesize queries by manipulating pairs of objects in the simulator: a stationary *anchor* and a *moving* object, which is related to the anchor by our concept. We investigate 9 spatial concepts:

1) *above*: angle between the objects' relative position and the world $z$-axis;
2) *above$_{bb}$*: intersection area of the two objects' bounding box projections on the world $xy$-plane;
3) *near*: inverse distance between the objects;
4) *upright*: angle between the moving object's $z$-axis and the world's;
5) *aligned$_{horiz}$*: angle between the objects' $x$-axes;
6) *aligned$_{vert}$*: angle between the objects' $z$-axes;
7) *forward*: angle between the anchor's $x$-axis and the objects' relative position;
8) *front*: angle between the anchor's $x$-axis and the objects' relative position;
9) *top*: angle between the anchor's $z$-axis and the objects' relative position.

For evaluation purposes, our ground truth concept implementations cut off the angles in *above*, *upright*, *aligned$_{horiz}$*, *aligned$_{vert}$*, *front*, and *top* after $45°$ and the distance in *near* after 0.3m, then normalize all concept values between 0 and 1. Fig. 2 showcases qualitative visualizations of the concepts.

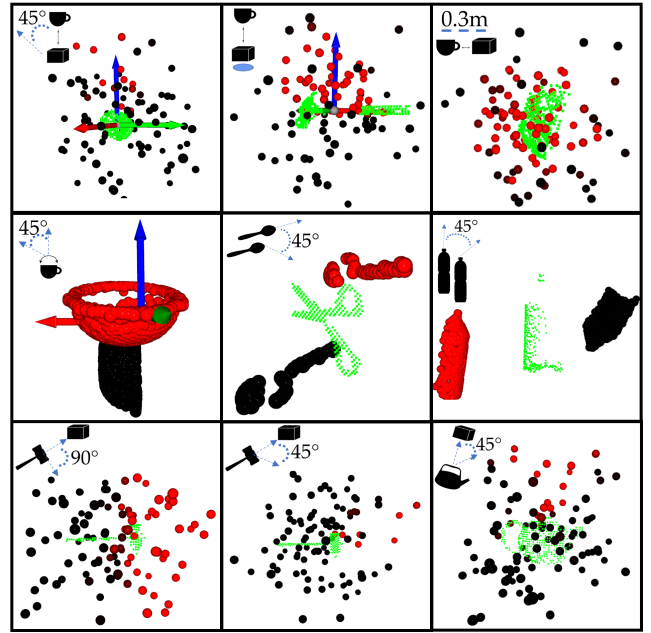

Fig. 2: Visual representation of the 9 perceptual concepts learned with our method (icon in the top left of each box). The anchor (green) is joined by examples of the moving object represented as either partial object point clouds (middle: *upright*, *aligned$_{horiz}$*, *aligned$_{vert}$*) or object point cloud centers (top: *above*, *above$_{bb}$*, *near*; bottom: *forward*, *front*, *top*). We color predicted positive examples in red, and negative ones in black. For concepts defined with respect to the world coordinate frame, we additionally plot the frame.

Notably, some of the concepts involve object affordances (*upright*, *aligned$_{horiz}$*, *aligned$_{vert}$*, *forward*, *front*, *top*). For those, only a subset of the objects are applicable (e.g. a mug has a *front*, but a box doesn't), so we selected object subsets for each concept accordingly (see App. VII-A). By default, the privileged space consists of the object poses, relative pose, positional difference, and bounding boxes.

We compare PCB to a baseline that learns $\phi_h$ directly from the human queries, without an intermediate low-dimensional concept. For PCB, we used the low-dimensional concepts $\phi_l$ trained using feature and label queries collected with the *confrand* and *augment* active learning strategies together. We show in Sec. V that this was the best performing $\phi_l$ with the overall cheapest type of human input, and results with other variants of $\phi_l$ follow similar trends (see App. VII-B). We use the concepts $\phi_l$ to label a set of 80,000 randomly generated training states, resulting in $\mathcal{D}_{\phi_l}$, then train $\phi_h$ using $\mathcal{D}_{\phi_l}$. Since the baseline is a PointNet that takes a long time to train, it is not suitable for active learning, so we train it with label queries generated with the *random* strategy. For additional comparison, we also train the baseline with the label queries collected by PCB with the *confrand* and *augment* strategies.

We train $\phi_h$ with each method and a varying number of queries, and report two metrics: 1) *Classification Accuracy*: how well the concepts can predict labels for a test set of states, and 2) *Optimization Accuracy*: how well the states induced by optimizing these concepts fare under the true $\phi^*$.

For *Classification Accuracy*, we use $\phi^*$ to generate a test

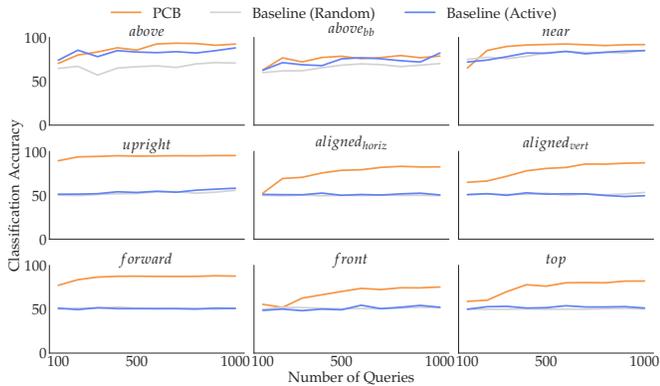

Fig. 3: Classification accuracy on a held-out test data set, for models trained on a varying number of queries. PCB concepts (orange) correctly classify at least 80% of the data after the first 500 queries in most cases. The baseline with random queries (gray) struggles to perform better than random; when trained with PCB's actively collected data (blue), the baseline performs better for the simple concepts but fails on the last six concepts that involve affordances.

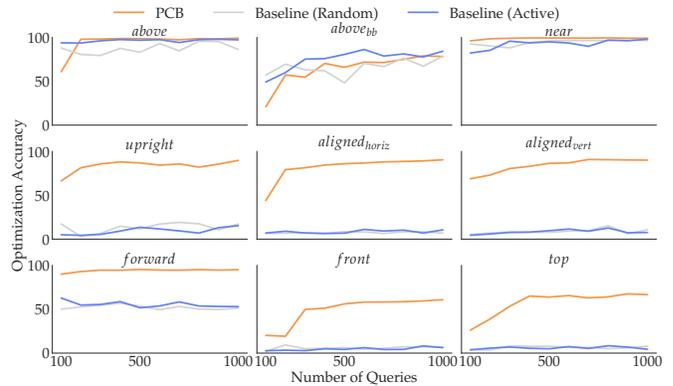

Fig. 4: Accuracy when optimizing object poses based on the learned concepts (*Optimization Accuracy*). PCB (orange) produces satisfactory poses for most concepts, as opposed to the baseline (gray and blue) which sometimes cannot even surpass 10% performance.

set $\mathcal{D}_{test}$ of 20,000 state-label pairs such that they have an equal number of positives and negatives. This way, we probe whether the learned concepts perform well on both labels and don't bias to one. We measure $\phi_h$'s accuracy as the percentage of datapoints in $\mathcal{D}_{test}$ predicted correctly.

For *Optimization Accuracy*, we sample 1,000 states $\mathcal{S}_{opt}$ with a concept value of 0, and use the learned concepts to optimize them into a new set of states $\mathcal{S}'_{opt}$. We do so by finding a pose transform on the moving object that maximizes the concept value, and use the cross-entropy method [29]. Importantly, since this is happening at test time, we use the high-dimensional observation of the state $o_h$ to perform the optimization. We evaluate $\mathcal{S}'_{opt}$ under $\phi^*$ and report the percentage of states that are labeled as 1. We present results for an arbitrarily chosen fixed seed.

### B. Qualitative results

Fig. 2 showcases the 9 concepts trained using PCB. For every concept, we show the anchor object in green (if applicable), together with positive and negative examples of the concept. For *above* and *above$_{bb}$* the positive examples above the anchor are sparse, which could make learning challenging from a data diversity perspective: if the robot doesn't query for enough positive examples, it won't be able to learn a meaningful decision boundary for these concepts. In contrast, *near* has a balanced mix of positives and negatives, making it a simpler concept to learn. The remaining six concepts all involve affordances which are dependent on the object shapes (e.g. a bowl is upright if its opening points upwards, an object is atop the kettle is it's placed above the lid, etc). Thus, to learn such concepts describing functionality across a plurality of objects and camera views, a method that learns directly from point clouds would need large amounts of data to accurately capture how the concepts relate to all objects' morphologies. The figure shows that PCB handles these challenges gracefully.

### C. Quantitative analysis

Fig. 3 shows *Classification Accuracy* results. Both variants of the baseline perform well for *above*, *above$_{bb}$*, and *near*, eventually reaching 70% performance. We think this happens because these concepts only require absolute position information, which is easy to infer from just the positions of the points clouds. The baseline trained on active data from PCB performs closer to PCB, suggesting that in some cases we may use the privileged information and low-dimensional concepts to guide the labeling process of high-dimensional data and be more sample efficient, without generating additional high-dimensional data. However, this does not always hold: the other six concepts involve affordances in addition to positions, which is much more challenging to capture with limited data. As a result, neither baseline version can achieve performance better than random. In contrast, PCB, which is able to generate thousands of high-dimensional training data points capturing different object point cloud morphologies, can successfully learn these kinds of concepts, correctly classifying at least 80% of the test data after the first 500 queries in most cases. In Fig. 4, *Optimization Accuracy* results tell a similar story. Our concepts can be optimized successfully with an accuracy of over 50%, meaning that we would be able to find positions for objects to satisfy these concepts [9]. Meanwhile, several baseline concepts have a success rate barely above 10%.

## V. ANALYSIS: LEARNING LOW-DIMENSIONAL CONCEPTS FROM DIFFERENT TYPES OF HUMAN QUERIES

In the previous section, we saw how our method, given a low-dimensional concept learned from human input, can outperform the baseline learning directly from high-dimensional sensor data. We now analyze what are the best strategies for learning low-dimensional concepts from human input. We seek to answer the following: **Q1:** Does querying via demonstration – the most informative type of query but also the most expensive – benefit learning when compared to random label queries? **Q2:** Does modifying the privileged information space via feature queries speed up learning? **Q3:** Can we choose label queries – the cheaper version of demo queries – that are more informative than random via active learning? **Q4:** How does labeling noise affect the quality of the learned concepts?

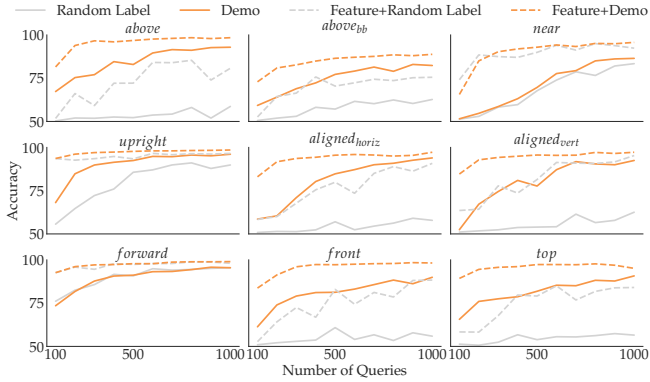

Fig. 5: Comparing different query and input space strategies. *Demo* queries outperform *random label* for concepts with few positives, and *feature* queries improve learning speed.

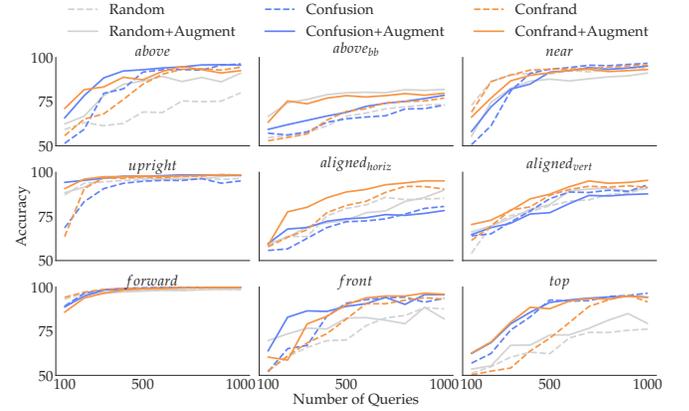

Fig. 6: Comparison amongst active labeling and positives selection strategies. *Confrand* is the most consistently beneficial strategy, and *Augment* boosts performance, especially in low data regimes.

### A. Benefits of Demonstration, Label, and Feature Queries

Our first experiment compares the three types of human queries across two dimensions: the query selection strategy and the privileged input space. For the former, while the robot could randomly synthesize states and ask the human to label them (i.e. random label queries), for some concepts such a strategy would rarely find states with positive concept values. In contrast, demonstration queries allow the human to select the states themselves, so they can balance the amount of positives and negatives the data set contains to be informative. As for the privileged input space, by default it contains many features that are correlated with one another or irrelevant to some concepts altogether. These redundant dimensions can make learning more difficult. Feature queries, with just a few simple and intuitive questions, can eliminate some dimensions of the input space that are unnecessary.

To answer **Q1** and **Q2**, we use a $2 \times 2$ factorial design. We manipulate the *query strategy* (*random label* and *demo*), and the *input space strategy* (*feature* and *no feature*). For both query strategies, we generate a dataset of labeled states as described in Sec. III-D, and simulate human input by sampling examples randomly for *random label* or in a way that balances positives and negatives for *demo*. The practical difference is in the positives-to-negatives ratio: while for *random label* that may be low for certain concepts (other than *near* and *forward* the mean ratio is 0.08), for *demo* it is 1. For *feature*, we ask three intuitive questions: F1. Does the concept concern a single object? F2. Does the concept care about the objects' absolute poses or their relative one? F3. Do the object sizes matter? F1 discards dimensions from the redundant object (useful for concepts like *upright*). F2 gets rid of correlated features (absolute or relative pose). F3 drops bounding box information if the concept doesn't require it.

We compare the learned concept network $\phi_l$ to the ground truth $\phi^*$ with a metric similar to *Classification Accuracy* from Sec. IV: we measure $\phi_l$'s correct prediction rate for $\mathcal{D}_{test}$.

In Fig. 5, we show results with varying amounts of queries from 100 to 1000. Comparing the solid lines, we immediately see that, for most concepts, demo queries perform much better than random label queries. The only concepts where this trend doesn't hold are *forward* and *near*, which are concepts where random sampling can already easily find many positives. This result stresses that having enough positives is crucial for learning good concepts. We can also compare the effect of feature queries: whether we use demo or label queries, feature queries considerably speed up learning, and this result holds across all 9 concepts. Another observation is that the combination of demo and feature queries plateaus in performance after about 200-300 queries, suggesting that, although each query requires more human time, the teaching process altogether might be shorter.

### B. Active Query Labeling

In Sec. V-A, we saw that demonstration queries substantially benefit concept learning when compared to random label queries. Unfortunately, demo queries are also very time-consuming to collect[1], which only makes them feasible in low-data regimes. In this section, we tackle **Q3** and explore whether we can make label queries more informative by employing active learning techniques, rather than simply randomly selecting them. This way, we can have the benefits of both informative query generation and easy label collection.

We use a $3 \times 2$ factorial design where we vary the *active strategy* (*random*, *confusion*, and *confrand*) and the *positives selection* (*augment* and *no augment*). As described in Sec. III, *random* generates a query state randomly and *confusion* picks a state at the decision boundary of the currently learned concept. We also introduce *confrand*, which randomly selects between the two strategies, to balance exploration of new areas and disambiguation of the current concept. With an *augment* positives selection, for every query the method also randomly chooses whether to exploit the space of positives it has found so far or just go with the selected active strategy. We use a batch size of 100. We train $\phi_l$ with each strategy and varying number of queries, and report accuracy on $\mathcal{D}_{test}$.

In Fig. 6, we show results with increasing number of queries across the 6 total label query selection strategies. Right off the bat, we see that active learning helps more the harder it is to find positives. For concepts like *forward* or *near*, random label queries do well because the positive-to-negative ratio is already high. For all other concepts, however, active learning helps considerably, certain techniques more than others. A general trend is that using *augment* queries outperforms not using them, especially in lower data regimes, confirming our intuition that finding positives earlier on improves learning. Amongst *random*, *confusion*, and *confrand*, we don't see a clear winner for all concepts, but *confrand*, the combination of novelty and uncertainty exploration, seems to perform the best across. It is encouraging to see that the performance can reach 80% accuracy after the first 500 queries, which would require a mere 10 minutes of human labeling time.

---

[1]Empirically, an expert user can label 100 queries in 2 minutes, but needs 10 minutes for the same amount of demo queries.

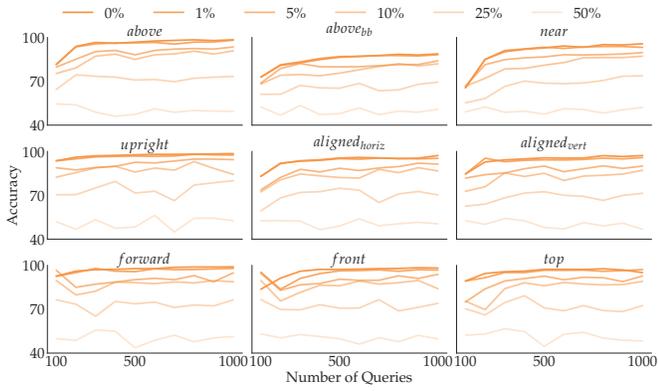

Fig. 7: Comparison for different labeling noise levels. Our method can withstand reasonable noise levels around 1-10%

### C. Noise Ablation in Human Query Labeling

Until now, we assumed the simulated human answered the queries perfectly. As this is not necessarily going to be the case for novice users, we examine **Q4**, how labeling noise affects our concept learning results. We do this by varying the *noise level* by 0%, 1%, 5%, 10%, 25%, and 50%. A "noisy" query has its label flipped. 50% is equivalent to a random labeler. We train $\phi_l$ by adding varying noise levels to the queries and report accuracy on $\mathcal{D}_{test}$. Fig. 7 reveals that, unsurprisingly, the noisier the queries, the worse the learned concept performs. While unrealistic noise levels like 25% or 50% severely worsen the quality of the learned concepts, our method seems to be able to withstand lower noise levels.

## VI. USING CONCEPTS IN MOTION PLANNING TASKS

We test our learned PCB concepts $\phi_h$ on a Franka Panda robot with a RGB-D camera in motion planning tasks, as shown in Fig. 1. For each trial, a user selects the concept to test, and the anchor and moving objects. Given the initial object placement, the robot's goal is to compute a new pose for the moving object that maximizes the concept value with respect to the anchor, grasp the object, and move it to the optimized pose. We used unknown object instance segmentation [36] to segment out the objects, 6-DOF GraspNet [35] to generate grasps, and RRT-connect [37] for motion planning.

We optimize a pose such that when applied to the moving object's (mean-centered) point cloud it results in a scene point cloud that maximizes the concept value. We use the Cross-Entropy Method (CEM) [29] with the cost function given by the concept value of the moved scene point cloud. To encourage the model to find object poses at reasonable orientations, we added a quaternion-angle cost to the CEM optimization, similarly to the metric used in prior work [38]:

$$d(q_1, q_2) = \lambda \left(1 - \langle q_1, q_2 \rangle\right)$$

where $q_1$ is the pose being optimized, $q_2 = I = (0, 0, 0, 1)$ is an identity quaternion, and $\lambda = 0.001$ is manually-tuned.

When optimized, our concept models found good poses, even on real-world data of previously unseen objects. Since our method's performance depends on that of the off-the-shelf point-cloud segmentation model used, failures may occur when large portions of the objects are occluded. Moreover, we focus on generating poses that maximize the concept without any collision constraints, so the moved objects sometimes collide with the anchor. In the future, we would fix this by incorporating these concepts into a planner such as that proposed in [9], so as to include the robot's kinematic constraints directly in the optimization process.

## VII. DISCUSSION AND CONCLUSIONS

In this paper, we presented a method for learning relational concepts with as little expert human interaction is possible. Our approach quickly learns a concept in a low-dimensional space, which is used to generate a large data set for training in a high-dimensional space such as the robot's sensor space.

While our results demonstrate that our concepts can be used on a 7-DoF Franka Panda arm operating with real sensor data, we still need to investigate how concepts taught by real people would fare. Our noise analysis in Sec. V-C suggests that limited random labeling noise might not affect the results too much, but this type of noise might not be a good model for how people make labeling errors. It could also be interesting to study the trade-off between learning accuracy and human burden for different types of queries.

Additionally, while we demonstrated our method in the context of object relations for manipulation, we are excited about future extensions to other types of concepts: many-object concepts ("the cup is *surrounded* by plates"), ordering ("sorted from largest to smallest"), or even functional relationships (support / concealment). We could extend PCB to any concepts where privileged information is available at training time. For instance, if the privileged space contains poses between two frames, we could learn an acceptable speed threshold for manipulating objects. A potential limitation is that the privileged space does become more complex the more (possibly correlated or irrelevant) information we add to it, so learning low-dimensional concepts may require more data. Our results give some evidence that PCB would overall still require much less data than learning directly from high dimensions, but more future work would be beneficial.

Finally, it could be worthwhile to study modifications to our training and query collection procedure to further improve the quality of the data. For example, we could combine demo and label queries by "warm-starting" the model with demo queries and then actively asking for label queries. The robot could also display examples of the currently learned concept to assist the person in deciding what new examples to give. Lastly, we could consider "chaining" learned concepts ("mug *upright* and in *front* of the hammer").

## References

[1] O. Mees, N. Abdo, M. Mazuran, and W. Burgard, "Metric learning for generalizing spatial relations to new objects," in *2017 IEEE/RSJ International Conference on Intelligent Robots and Systems (IROS)*, 2017, pp. 3175–3182.

[2] A. Thippur, C. Burbridge, L. Kunze, M. Alberti, J. Folkesson, P. Jensfelt, and N. Hawes, "A comparison of qualitative and metric spatial relation models for scene understanding," in *Proceedings of the Twenty-Ninth AAAI Conference on Artificial Intelligence*, ser. AAAI'15. AAAI Press, 2015, p. 1632–1640.

[3] T. Mota and M. Sridharan, "Incrementally grounding expressions for spatial relations between objects," in *Proceedings of the 27th International Joint Conference on Artificial Intelligence*, ser. IJCAI'18. AAAI Press, 2018, p. 1928–1934.

[4] Y. Xiang, T. Schmidt, V. Narayanan, and D. Fox, "Posecnn: A convolutional neural network for 6d object pose estimation in cluttered scenes," 06 2018.

[5] M. Sundermeyer, Z.-C. Márton, M. Durner, M. Brucker, and R. Triebel, "Implicit 3d orientation learning for 6d object detection from rgb images," in *ECCV*, 2018.

[6] X. Deng, A. Mousavian, Y. Xiang, F. Xia, T. Bretl, and D. Fox, "Poserbpf: A rao–blackwellized particle filter for 6-d object pose tracking," *IEEE Transactions on Robotics*, vol. 37, no. 5, pp. 1328–1342, 2021.

[7] C. Wang, R. Martín-Martín, D. Xu, J. Lv, C. Lu, L. Fei-Fei, S. Savarese, and Y. Zhu, "6-pack: Category-level 6d pose tracker with anchor-based keypoints," in *2020 IEEE International Conference on Robotics and Automation (ICRA)*, 2020, pp. 10 059–10 066.

[8] K. Kase, C. Paxton, H. Mazhar, T. Ogata, and D. Fox, "Transferable task execution from pixels through deep planning domain learning," *2020 IEEE International Conference on Robotics and Automation (ICRA)*, pp. 10 459–10 465, 2020.

[9] C. Paxton, C. Xie, T. Hermans, and D. Fox, "Predicting stable configurations for semantic placement of novel objects," in *Conference on Robot Learning (CoRL)*, 2021, to appear.

[10] W. Yuan, C. Paxton, K. Desingh, and D. Fox, "Sornet: Spatial object-centric representations for sequential manipulation," 2021.

[11] S. Mukherjee, C. Paxton, A. Mousavian, A. Fishman, M. Likhachev, and D. Fox, "Reactive long horizon task execution via visual skill and precondition models," in *2021 IEEE/RSJ International Conference on Intelligent Robots and Systems (IROS)*. IEEE, 2020, pp. 5717–5724.

[12] B. D. Ziebart, A. Maas, J. A. Bagnell, and A. K. Dey, "Maximum entropy inverse reinforcement learning," in *Proceedings of the 23rd National Conference on Artificial Intelligence - Volume 3*, ser. AAAI'08. AAAI Press, 2008, pp. 1433–1438.

[13] P. Abbeel and A. Y. Ng, "Apprenticeship learning via inverse reinforcement learning," in *Machine Learning (ICML), International Conference on*. ACM, 2004.

[14] D. Hadfield-Menell, S. Milli, P. Abbeel, S. J. Russell, and A. Dragan, "Inverse reward design," in *Advances in Neural Information Processing Systems*, I. Guyon, U. V. Luxburg, S. Bengio, H. Wallach, R. Fergus, S. Vishwanathan, and R. Garnett, Eds., vol. 30. Curran Associates, Inc., 2017.

[15] J. Choi and K.-E. Kim, "Bayesian nonparametric feature construction for inverse reinforcement learning," in *Twenty-Third International Joint Conference on Artificial Intelligence*, 2013.

[16] S. Levine, Z. Popovic, and V. Koltun, "Feature construction for inverse reinforcement learning," in *Advances in Neural Information Processing Systems*, 2010, pp. 1342–1350.

[17] A. Bobu, M. Wiggert, C. Tomlin, and A. D. Dragan, "Feature expansive reward learning: Rethinking human input," in *Proceedings of the 2021 ACM/IEEE International Conference on Human-Robot Interaction*, ser. HRI '21. New York, NY, USA: Association for Computing Machinery, 2021, p. 216–224.

[18] A. Bobu, M. Wiggert, C. J. Tomlin, and A. D. Dragan, "Inducing structure in reward learning by learning features," *CoRR*, vol. abs/2201.07082, 2022.

[19] C. Finn, S. Levine, and P. Abbeel, "Guided cost learning: Deep inverse optimal control via policy optimization," in *Proceedings of the 33rd International Conference on International Conference on Machine Learning - Volume 48*, ser. ICML'16. JMLR.org, 2016, p. 49–58.

[20] M. Wulfmeier, D. Z. Wang, and I. Posner, "Watch this: Scalable cost-function learning for path planning in urban environments," in *2016 IEEE/RSJ International Conference on Intelligent Robots and Systems (IROS)*, 2016, pp. 2089–2095.

[21] L. Shao, T. Migimatsu, Q. Zhang, K. Yang, and J. Bohg, "Concept2robot: Learning manipulation concepts from instructions and human demonstrations," in *Robotics: Science and Systems*, 2020.

[22] J. Fu, K. Luo, and S. Levine, "Learning robust rewards with adverserial inverse reinforcement learning," in *International Conference on Learning Representations*, 2018.

[23] S. Reddy, A. D. Dragan, and S. Levine, "SQIL: imitation learning via reinforcement learning with sparse rewards," in *8th International Conference on Learning Representations, ICLR 2020, Addis Ababa, Ethiopia, April 26-30, 2020*. OpenReview.net, 2020.

[24] Y. Hristov, D. Angelov, M. Burke, A. Lascarides, and S. Ramamoorthy, "Disentangled relational representations for explaining and learning from demonstration," in *3rd Annual Conference on Robot Learning, CoRL 2019, Osaka, Japan, October 30 - November 1, 2019, Proceedings*, ser. Proceedings of Machine Learning Research, L. P. Kaelbling, D. Kragic, and K. Sugiura, Eds., vol. 100. PMLR, 2019, pp. 870–884.

[25] A. Goyal, A. Mousavian, C. Paxton, Y.-W. Chao, B. Okorn, J. Deng, and D. Fox, "Ifor: Iterative flow minimization for robotic object rearrangement," 2022.

[26] M. Cakmak and A. L. Thomaz, "Designing robot learners that ask good questions," in *2012 7th ACM/IEEE International Conference on Human-Robot Interaction (HRI)*, 2012, pp. 17–24.

[27] S. Reddy, A. Dragan, S. Levine, S. Legg, and J. Leike, "Learning human objectives by evaluating hypothetical behavior," in *ICML*, 2020.

[28] E. Biyik, K. Wang, N. Anari, and D. Sadigh, "Batch active learning using determinantal point processes," *CoRR*, vol. abs/1906.07975, 2019.

[29] P.-T. De Boer, D. P. Kroese, S. Mannor, and R. Y. Rubinstein, "A tutorial on the cross-entropy method," *Annals of operations research*, vol. 134, no. 1, pp. 19–67, 2005.

[30] M. Kobilarov, "Cross-entropy randomized motion planning," 06 2011.

[31] E. Wijmans, "Pointnet++ pytorch," *https://github.com/erikwijmans/Pointnet2_PyTorch*, 2018.

[32] C. R. Qi, L. Yi, H. Su, and L. J. Guibas, "Pointnet++: Deep hierarchical feature learning on point sets in a metric space," in *Advances in Neural Information Processing Systems*, 2017, pp. 5099–5108.

[33] A. X. Chang, T. Funkhouser, L. Guibas, P. Hanrahan, Q. Huang, Z. Li, S. Savarese, M. Savva, S. Song, H. Su, J. Xiao, L. Yi, and F. Yu, "ShapeNet: An Information-Rich 3D Model Repository," Stanford University — Princeton University — Toyota Technological Institute at Chicago, Tech. Rep. arXiv:1512.03012 [cs.GR], 2015.

[34] V. Makoviychuk, L. Wawrzyniak, Y. Guo, M. Lu, K. Storey, M. Macklin, D. Hoeller, N. Rudin, A. Allshire, A. Handa, and G. State, "Isaac gym: High performance gpu-based physics simulation for robot learning," 2021.

[35] A. Mousavian, C. Eppner, and D. Fox, "6-DOF graspnet: Variational grasp generation for object manipulation," in *International Conference on Computer Vision (ICCV)*, 2019.

[36] Y. Xiang, C. Xie, A. Mousavian, and D. Fox, "Learning rgb-d feature embeddings for unseen object instance segmentation," in *Conference on Robot Learning (CoRL)*, 2020.

[37] J. Kuffner and S. LaValle, "Rrt-connect: An efficient approach to single-query path planning," in *Proceedings 2000 ICRA. Millennium Conference. IEEE International Conference on Robotics and Automation. Symposia Proceedings (Cat. No.00CH37065)*, vol. 2, 2000, pp. 995–1001 vol.2.

[38] W. Yang, C. Paxton, A. Mousavian, Y.-W. Chao, M. Cakmak, and D. Fox, "Reactive human-to-robot handovers of arbitrary objects," in *IEEE International Conference on Robotics and Automation (ICRA)*, 2021.

# APPENDIX

## A. Concept Objects

For data generation, we modified the objects in the ShapeNet dataset [33] such that they are consistently aligned and scaled. We selected objects commonly found in tabletop manipulation tasks, like bowls, cereal boxes, cups, cans, mugs, bottles, cutlery, hammers, candles, teapots, fruit, etc. (see Fig. 8). The concepts *above*, *above$_{bb}$*, and *near* used all the selected objects because they don't involve object affordances. For the concepts that involve affordances, we selected subsets from the object set accordingly. For *upright* and *top*, we used objects with evident upright orientations: bottles, bowls, candles, mugs, cups, cans, milk cartons, pans, plates, and teapots. For *aligned$_{horiz}$* we used objects that can be horizontally aligned: calculators, can openers, cutlery, hammers, pans, and scissors. For *aligned$_{vert}$* we used objects that can be vertically aligned: bottles, boxes, candles, cups, milk cartons, and cans. For *forward* and *front* we used large enough objects with clear fronts: hammers, pans, and teapots.

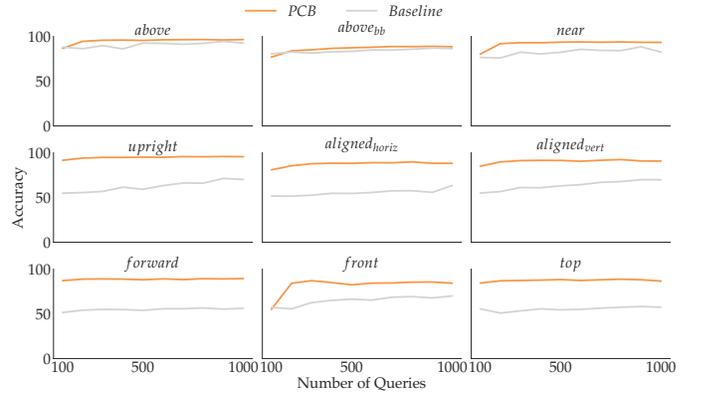

Fig. 9: Classification accuracy on a held-out test data set (*Classification Accuracy*), for models trained on a varying number of queries. Concepts trained using our PCB method (orange) correctly classify at least 80% of the test data after the first 200 demo queries. Meanwhile, the baseline (gray) struggles to perform better than random, especially on the last six concepts that involve affordances.

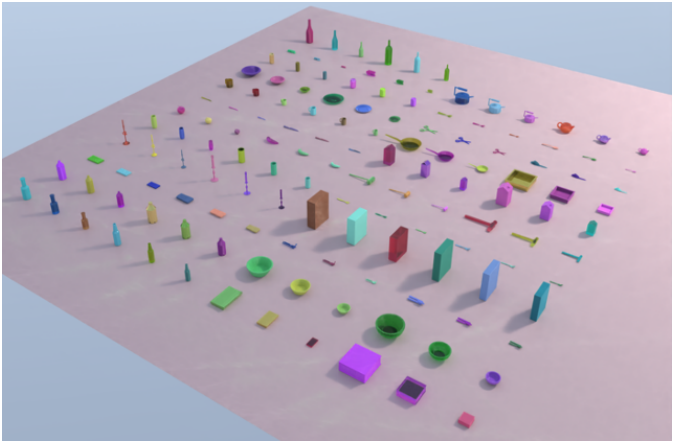

Fig. 8: We show the well-aligned and scaled ShapeNet objects we used. We chose objects commonly found in manipulation tasks.

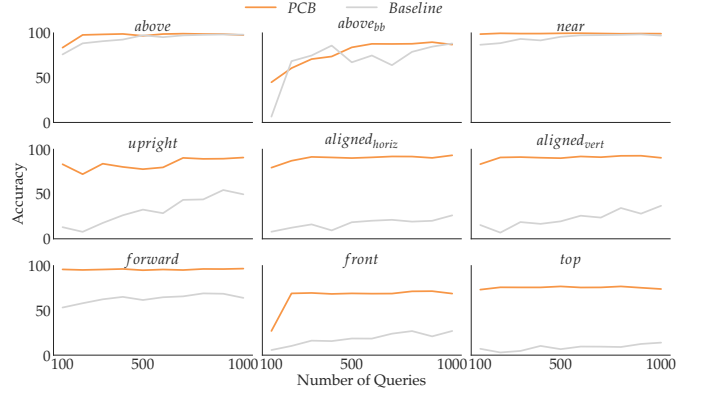

Fig. 10: Accuracy when optimizing object poses based on the learned concepts (*Optimization Accuracy*). Our PCB method (orange) produces satisfactory poses for most concepts, as opposed to the baseline (gray) which sometimes cannot even surpass 25% performance.

## B. PCB Results for Demo Queries

In this section, we expand on the results in Sec. IV by showing the case where the human provides the robot with demonstration queries. We compare PCB to a baseline that learns $\phi_h$ directly from the queries. For PCB, we take the $\phi_l$ concepts we trained using both demonstration and feature queries in Sec. V-A, and use them to label a large set of 80,000 training states, resulting in $\mathcal{D}_{\phi_l}$. Our method then trains $\phi_h$ using $\mathcal{D}_{\phi_l}$, while the baseline trains the same architecture using the original queries we used to learn $\phi_l$. Importantly, both methods use well-balanced demonstration queries. We report results on the same two metrics from Sec. IV, *Classification Accuracy* and *Optimization Accuracy*.

Fig. 9 shows *Classification Accuracy* results. The baseline actually performs well for *above*, *above$_{bb}$*, and *near*, eventually reaching 80% performance. We think this happens because for these concepts it is easy to infer the necessary privileged information just from the positions of the point clouds. For example, for *near*, given the position of the two object point cloud centers, learning a relationship between their distance and the concept value should not require more than a few samples. The other concepts involve affordances in addition to position information, which is much more challenging to capture. As a result, the baseline can barely achieve performance better than random. In contrast, our method, which is able to generate thousands of high-dimensional training data points, can successfully learn these kinds of concepts, correctly classifying at least 80% of the test data after the first 200 queries. Note that PCB with demo queries reaches this accuracy faster than with the label queries from Sec. IV, but demo queries are more effortful to give than label queries. This shows the trade-off between human effort and informativeness we investigated in Sec. V.

In Fig. 4, *Optimization Accuracy* results tell a similar story. Our concepts can be optimized successfully with an accuracy of over 50%, meaning that we would be able to find positions for objects to satisfy these concepts [9]. Meanwhile, several baseline concepts have a success rate barely above 25%.